\documentclass[conference,a4paper]{IEEEtran}

\usepackage[utf8]{inputenc} 
\usepackage[T1]{fontenc}
\usepackage{hyperref} 
\hypersetup{
    pdfnewwindow=true,     
    colorlinks=true,     
    linkcolor=blue,     
    citecolor=red,     
    filecolor=magenta,     
    urlcolor=cyan     
}
\usepackage{url}              
\usepackage{cite}             

\usepackage[cmex10]{amsmath}  
\usepackage{mleftright}       
\mleftright                   

\usepackage{graphicx}         
\usepackage{booktabs}         
\usepackage{amssymb}
\usepackage{amsfonts}
\usepackage{tikz}

\usepackage{amsmath,amsfonts,bm}

\def\1{\bm{1}}

\def\vmu{{\bm{\mu}}}

\def\vu{{\bm{u}}}

\def\vw{{\bm{w}}}

\def\vz{{\bm{z}}}

\def\mA{{\bm{A}}}
\def\mB{{\bm{B}}}

\def\mI{{\bm{I}}}

\def\mN{{\bm{N}}}

\def\mX{{\bm{X}}}
\def\mY{{\bm{Y}}}
\def\mZ{{\bm{Z}}}

\def\mSigma{{\bm{\Sigma}}}

\DeclareMathAlphabet{\mathsfit}{\encodingdefault}{\sfdefault}{m}{sl}
\SetMathAlphabet{\mathsfit}{bold}{\encodingdefault}{\sfdefault}{bx}{n}

\newtheorem{theorem}{Theorem}
\newtheorem{lemma}{Lemma}

\newtheorem{remark}{Remark}

\newcommand{\Expt}{\mathbb{E}}
\newcommand{\nn}{\notag}
\newcommand{\diff}{\mathrm{d}}

\begin{document}

\title{On the Generalization Error of Meta Learning\\ for the Gibbs Algorithm} 


%
%

%
\author{%
  \IEEEauthorblockN{Yuheng Bu\IEEEauthorrefmark{1},
                    Harsha Vardhan Tetali\IEEEauthorrefmark{1},
                    Gholamali Aminian\IEEEauthorrefmark{2},
                    Miguel Rodrigues\IEEEauthorrefmark{3}
                    and Gregory Wornell\IEEEauthorrefmark{4}
                    }

  \IEEEauthorblockA{\IEEEauthorrefmark{1}%
                    University of Flroida,  \IEEEauthorrefmark{2}%
                    The Alan Turing Institute,  \IEEEauthorrefmark{3}%
                   University College London, \IEEEauthorrefmark{4}%
                    Massachusetts Institute of Technology}
  \IEEEauthorblockA{Email: \{buyuheng, vardhanh71\!\}@ufl.edu, gaminian@turing.ac.uk, m.rodrigues@ucl.ac.uk, gww@mit.edu}                  
}

\maketitle

\begin{abstract}
We analyze the generalization ability of joint-training meta learning algorithms via the Gibbs algorithm. Our exact characterization of the expected meta generalization error for the meta Gibbs algorithm is based on symmetrized KL information, which measures the dependence between all meta-training datasets and the output parameters, including task-specific and meta parameters. Additionally, we derive an exact characterization of the meta generalization error for the super-task Gibbs algorithm, in terms of conditional symmetrized KL information within the super-sample and super-task framework introduced in \cite{steinke2020reasoning} and \cite{hellstrom2022evaluated}, respectively. Our results also enable us to provide novel distribution-free generalization error upper bounds for these Gibbs algorithms applicable to meta learning.


\end{abstract}

\section{Introduction}

In meta learning problems,\footnote{a.k.a. lifelong learning or learning to learn} we have access to multiple related tasks generated from a task environment, and our goal is to capture the shared information among all tasks and construct a model that can generalize to new tasks drawn from the same environment. State-of-the-art meta learning algorithms---such as \cite{finn2017model}---have been successfully used in a wide range of applications, including object detection, data mining, few-shot learning, continual learning,  and natural language processing \cite{liu2020deep,brazdil2008metalearning,liu2021pre,obamuyide2019model,harrison2020continuous}.

Various analyses have been pursued to explain the success of meta learning. For example, \cite{baxter2000model} introduces the task environment concept in meta learning, and derives generalization upper bounds via uniform convergence. Other techniques, such as PAC-Bayesian and information-theoretic approaches, have been adopted to construct generalization error bounds, demonstrating both environment and task-level dependencies in the generalization behavior of meta learning. High probability PAC-Bayesian bounds have been proposed in \cite{pentina2014pac,amit2018meta,liu2021pac,rothfuss2021pacoh,rezazadeh2022general}. Inspired by \cite{russo2019much,xu2017information,bu2020tightening}, information-theoretic upper bounds on the expected generalization error of meta learning are developed in \cite{jose2021information}, and later refined in \cite{chen2021generalization}, which bounds the meta generalization error using mutual information for both joint-training\footnote{Meta and task-specific parameters are updated within the same dataset.} and alternate-training\footnote{Meta parameters and task-specific parameters are updated within two different datasets.} algorithms. More recently,  \cite{hellstrom2022evaluated} develops upper bounds on the meta generalization error in terms of evaluated conditional mutual information via a super-task framework, which extends the super-sample approach in \cite{steinke2020reasoning}. However, it is important to appreciate that such upper bounds may not fully capture the generalization ability of a meta learning algorithm, as the tightness of the bounds is subject to the limitations of the bounding technique.


In contrast to such approaches, we develop exact characterizations of the generalization errors for joint-training meta learning algorithms via the Gibbs algorithm. We model the empirical meta risk minimization algorithm proposed by \cite{chen2021generalization} via a meta Gibbs algorithm. We also consider a super-task Gibbs algorithm inspired by the super-task framework in \cite{hellstrom2022evaluated}. 

Our main contributions of this work are as follows:
\begin{itemize}
    \item We provide an exact characterization of the meta generalization error for the meta Gibbs algorithm in terms of symmetrized KL information.   
    \item We provide an exact characterization of the meta generalization error for the Gibbs algorithm in super-task framework \cite{hellstrom2022evaluated} using conditional symmetrized KL information.
    \item Using our exact characterizations of the meta generalization error, we provide distribution-free upper bounds, which expose the convergence rate of the meta generalization error of the joint-training Gibbs algorithms in terms of the number of samples and tasks.
\end{itemize}

\vspace{-0.5em}

\section{Preliminaries}
Our exact characterizations involve various information measures. If $P$ and $Q$ are probability measures over space $\mathcal{X}$, and $P$ is absolutely continuous with respect to $Q$, the Kullback-Leibler (KL) divergence between $P$ and $Q$ is given by
$D(P\|Q)\triangleq\int_\mathcal{X}\log\left(\frac{\diff P}{\diff Q}\right) \diff P$. If $Q$ is also absolutely continuous with respect to $P$, the symmetrized KL divergence (i.e., Jeffrey's divergence~\cite{jeffreys1946invariant}) is
\begin{equation}
D_{\mathrm{SKL}}(P\|Q)\triangleq D(P \| Q) + D(Q\|P).
\end{equation}
The mutual information between random variables $X$ and $Y$ is the KL divergence between the joint distribution and product-of-marginal
distribution $I(X;Y)\triangleq D(P_{X,Y}\|P_X\otimes P_{Y})$. 
Swapping the role of $P_{X,Y}$ and $P_X\otimes P_{Y}$ in mutual information, we obtain the lautum information introduced by \cite{palomar2008lautum},
\begin{equation}
    L(X;Y)\triangleq D(P_X\otimes P_{Y}\| P_{X,Y}).
\end{equation}
The symmetrized KL information \cite{aminian2015capacity} between $X$ and $Y$ is
\begin{align*}
   I_{\mathrm{SKL}}(X;Y)\!\triangleq\! D_{\mathrm{SKL}}(P_{X,Y}\|P_X\otimes P_Y)\!=\!
   I(X;Y)+ L(X;Y).
\end{align*}
The conditional mutual information between two random variables $X$ and $Y$ conditioned on $Z$ is the KL divergence between $P_{X,Y|Z}$ and $P_{X|Z} \otimes P_{Y|Z}$ averaged over $P_{Z}$, \begin{align*}
    I(X;Y|Z)\triangleq \mathbb{E}_{P_Z}[D(P_{X,Y|Z=z} \| P_{Y|Z=z} \otimes P_{X|Z=z})].
\end{align*}
Similarly, we can also define the conditional lautum information $L(X;Y|Z)$ and the conditional symmetrized KL information 
\begin{align}
   &I_{\mathrm{SKL}}(X;Y|Z)\triangleq
   I(X;Y|Z)+ L(X;Y|Z).
\end{align}

The $(\gamma,\pi(y),f(y,x))$-Gibbs distribution (a.k.a. Gibbs posterior~\cite{catoni2007pac}), which was first proposed by \cite{gibbs1902elementary} in statistical mechanics and further investigated by \cite{jaynes1957information} in information theory, is defined as:
\begin{equation}\label{Eq: Gibbs Solution}
    P_{Y|X}^\gamma (y|x) \triangleq \frac{\pi({y})\, e^{-\gamma f(y,x)}}{V_f(x,\gamma)},\quad \gamma\ge 0,
\end{equation}
where $\gamma$ is the inverse temperature, $\pi(y)$ is a prior distribution on $\mathcal{Y}$, $f(y,x)$ is energy function, and 
\begin{equation*}
V_f(x,\gamma) \triangleq \int \pi(y) e^{-\gamma f(y,x)} \diff y
\end{equation*}
 is the partition function.

\section{Background and Related Work}

\textbf{Motivations for Gibbs Algorithm:} In supervised learning, the Gibbs algorithm can be viewed as a \emph{randomized} empirical risk minimization (ERM) algorithm. In addition, the Stochastic Gradient Langevin Dynamics (SGLD) algorithm is known to converge to the Gibbs algorithm \cite{raginsky2017non}. The Gibbs algorithm can also be interpreted as the solution to the KL-divergence-regularized ERM problem \cite{xu2017information,zhang2006information,zhang2006E}. For more detailed discussions of the Gibbs algorithm, see, e.g., \cite{OurGibbsPaper}.

\textbf{Gibbs Algorithm and Generalization Error:}
An exact characterization of the generalization error of the Gibbs algorithm in terms of symmetrized KL information is provided in \cite{OurGibbsPaper} for supervised learning. 
The authors also provide a generalization error upper bound with the rate of $\mathcal{O}\left(1/n\right)$ under the sub-Gaussian assumption, where $n$ is the number of training samples. An information-theoretic upper bound with a similar  $\mathcal{O}\left(1/n\right)$ rate is provided by \cite{raginsky2016information} for the Gibbs algorithm with bounded loss function, and PAC-Bayesian bounds using a variational approximation of Gibbs posteriors are studied by \cite{alquier2016properties}. Both \cite{asadi2020chaining,kuzborskij2019distribution} focus on bounding the excess risk of the Gibbs algorithm in supervised learning. The generalization errors of the Gibbs algorithm in transfer learning and semi-supervised learning settings have been analyzed in \cite{bu2022characterizing} and \cite{he2023does}, respectively.

\textbf{Other Analysis of Meta Learning:} Besides the information-theoretic approach to analyze generalization error, there are other analyses of meta learning. For example, the uniform convergence analysis of meta learning is first conducted in \cite{baxter2000model}, and \cite{maurer2005algorithmic} adopts the tool of algorithmic stability. Distribution-dependent lower bounds on the meta learning algorithms are provided in \cite{konobeev2021distribution}.

\section{Meta Generalization Error of the Meta Gibbs Algorithm}

\subsection{Problem Formulation}
In meta learning, we aim to learn a model from multiple meta-training tasks that generalize to an unseen new task. Following \cite{baxter2000model,chen2021generalization},
we assume that all tasks are generated from a common environment $\tau$ with a meta distribution $P_\tau$ over the probability measures defined on $\mathcal{Z}$ as the space of data samples. We denote $m$ different meta-training tasks i.i.d. drawn from the meta distribution as $M_i \sim P_\tau$, $i\in [m]$. Without loss of generality, we assume that there are $n$ training samples $D_{M_i}=\{Z_j^{M_i}\}_{j=1}^n$ for each meta-training task $M_i$, which are generated (not necessarily i.i.d.) from the source distribution $P_{D_{M_i}}$.

As all tasks, including the unseen test task, are generated from the same meta distribution $P_\tau$, we can use a meta parameter $U \in \mathcal{U}$ to capture the shared knowledge among all tasks and  $W_{1:m}=(W_1,\cdots,W_{m})$ to denote the task specific-parameters. Here, we adopt a similar formulation as in the two-stage transfer learning considered by \cite{bu2022characterizing}, where the performance of $(U,W_i)$ is measured by a non-negative loss function $\ell: \mathcal{U} \times \mathcal{W} \times \mathcal{Z} \to \mathbb{R}_0^+$.
Thus, we define the following individual empirical risk for a single meta-training task $M_i$
\begin{align}
    L_E(U,W_i,D_{M_i})\triangleq\frac{1}{n}\sum_{j=1}^{n}\ell(U,W_i,Z_j^{M_i}),
\end{align}
and the joint empirical risk for all meta-training tasks
\begin{align}
    L_E(U,W_{1:m},D_{M_{1:m}})\triangleq\frac{1}{m}\sum_{i=1}^{m}L_E(U,W_i,D_{M_i}).
\end{align}

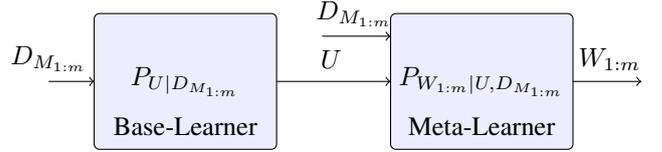
\begin{figure}[t]
\centering
\begin{tikzpicture}[scale=0.6]

\draw[->] (-1.5,5.5) -- (-0.5,5.5);
\node at (-1.5,6){$D_{M_{1:m}}$};

\draw[->] (3,5.5) -- (6,5.5);
\draw[->] (4.5,6.5) -- (6,6.5);


\draw [fill=white!93!blue,rounded corners=2pt](-0.5,4) rectangle (3.5,7);
\node at (1.5,5.5){$P_{U|D_{M_{1:m}}}$};
\node at (1.5,4.5){Base-Learner};

 \node at (4.7,6){$U$};
\node at (5.2,7){$D_{M_{1:m}}$};

\draw [fill=white!93!blue,rounded corners=2pt](6,4) rectangle (10,7);
\node at (8,5.5){$P_{W_{1:m}|U,D_{M_{1:m}}}$};
\node at (8,4.5){Meta-Learner};
\node at (10.8,6){$W_{1:m}$};
\draw[->] (10,5.5) -- (11.5,5.5);
\end{tikzpicture}
\caption{Joint-training meta learning algorithm}
\label{Fig: Meta}
\vspace{-1em}
\end{figure}

A meta learning algorithm, shown in Figure~\ref{Fig: Meta}, can be decomposed into two components, i.e., a \textit{meta-learner} and a \textit{base-learner}. The meta-learner maps all the dataset of training tasks to a random meta parameter $P_{U|D_{M_{1:m}}}$, and the base-learner maps the meta parameter and dataset of each task to specific parameters, i.e., $P_{W_{1:m}|U,D_{M_{1:m}}}\!\!= \prod_{i=1}^n P_{W_{i}|U,D_{M_{i}}}$.

We focus on the joint-training meta learning algorithm defined in \cite{chen2021generalization}.  In a joint-training algorithm, the training dataset $D_{M_{1:m}}$ are used to obtain all the task-specific parameters $W_{1:m}$ and meta parameter $U$ jointly,
which gives the following definition of \textit{empirical meta risk} for meta parameter $U$,
\begin{align}
    L_E (U,D_{M_{1:m}})\triangleq\frac{1}{m}\sum_{i=1}^{m}\mathbb{E}_{P_{W_{i}|U,D_{M_i}}}\left[L_E(U,W_i,D_{M_i})\right].
\end{align}

To evaluate the quality of the meta parameter $U$, an unseen test task $T$ is drawn from the environment $\tau$ with distribution $P_{\tau}$. We now define the \textit{population meta risk} as follows,
\begin{align}
    L_P (U,\tau)\!\triangleq\mathbb{E}_{P_\tau} \!\big[\mathbb{E}_{P_{D_{T}}}[\mathbb{E}_{P_{W_T|U,D_{T}}}[L_P(U,W_T,P_{D_{T}})]]\big],
\end{align}
where $D_{T}$ contains $n$ samples drawn from the test task $T$, and
$L_P(U,W,P_{D}) = \mathbb{E}_{P_{D} }[L_E(U,W,D)]$ denotes the  standard population risk.

Finally, the expected \emph{meta generalization error} that quantifies the generalizability of the meta parameter $U$ for the meta learning is
\begin{align}\label{Eq: expected GE meta}
&\overline{\text{gen}}(P_{W_{1:m}|U,D_{M_{1:m}}},P_{U|D_{M_{1:m}}},\tau) \nn \\
&\quad\triangleq \mathbb{E}_{P_\tau} \big[ \mathbb{E}_{P_{U,D_{M_{1:m}}}}[ L_P (U,\tau)- L_E (U,D_{M_{1:m}})]\big].
\end{align}

To understand the generalization error in meta learning, we consider the following meta Gibbs algorithm, i.e., $(\gamma,\pi(u, w_{1:m}),L_E(u,w_{1:m},d_{M_{1:m}}))$-Gibbs algorithm,
\begin{align}\label{Eq: meta Gibbs algorithm}
   & P_{W_{1:m},U|D_{M_{1:m}}}^\gamma (w_{1:m},u|d_{M_{1:m}}) \nn\\
   &\quad = \frac{\pi(u, w_{1:m}) e^{-\gamma L_E(u,w_{1:m},d_{M_{1:m}})}}{V(d_{M_{1:m}},\gamma)}.
\end{align}
 Note that this meta Gibbs algorithm is defined by learning $U$ and $W_{1:m}$ jointly. Due to the structure in the joint empirical risk $L_E(U,W_{1:m},D_{M_{1:m}})$, it can be verified that the induced base-learner satisfies the condition $P_{W_{1:m}|U,D_{M_{1:m}}} = \prod_{i=1}^n P_{W_{i}|U,D_{M_{i}}}$, i.e.,  $W_i$ only depends on $D_{M_{i}}$ conditioning on the meta-parameter $U$.

\subsection{Characterization of Expected Meta Generalization Error}
The following theorem provides an exact characterization of the expected meta generalization error of the meta Gibbs algorithm using symmetrized KL information. The proof is provided in Appendix~\ref{app:thm1}.
%
%
\begin{theorem}\label{Theorem: Meta-Gibbs} 
For the meta Gibbs algorithm in~\eqref{Eq: meta Gibbs algorithm}, the expected meta generalization error is
\begin{equation*}
    \overline{\text{gen}}(P_{W_{1:m},U|D_{M_{1:m}}}^\gamma,\tau)\! =\! \frac{\mathbb{E}_{P_\tau}[I_{\mathrm{SKL}}(U,W_{1:m};D_{M_{1:m}})]}{\gamma }.
\end{equation*}
\end{theorem}

Theorem~\ref{Theorem: Meta-Gibbs} only assumes that the meta-training tasks $P_{D_{M_i}}$ are i.i.d generated from $P_\tau$, and it holds even when the $n$ samples in $D_{M_i}=\{Z_j^{M_i}\}_{j=1}^n$ are not i.i.d.

Some basic properties of the expected meta generalization error can be proved directly from the properties of symmetrized KL information. 

\paragraph{Non-negativity} The non-negativity of the expected meta generalization error, i.e., $\overline{\text{gen}}(P_{W_{1:m},U|D_{M_{1:m}}}^\gamma,\tau)\ge 0$, follows from the non-negativity of $I_{\mathrm{SKL}}(U,W_{1:m};D_{M_{1:m}})$. 

\paragraph{Concavity} It is shown in \cite{aminian2015capacity} that the symmetrized KL information $I_{\mathrm{SKL}}(X;Y)$ is a concave function of $P_X$ for fixed $P_{Y|X}$. Thus, we have 
\begin{align}
    &{\mathbb{E}_{P_\tau}[I_{\mathrm{SKL}}(P_{W_{1:m},U|D_{M_{1:m}}}^\gamma\!\!, P_{D_{M_{1:m}}})]} \nn\\
    &\quad\le I_{\mathrm{SKL}}(P_{W_{1:m},U|D_{M_{1:m}}}^\gamma\!\!, \mathbb{E}_{P_\tau}[P_{D_{M_{1:m}}}]).
\end{align}
Note that $\mathbb{E}_{P_\tau}[P_{D_{M_{1:m}}}]$ can be viewed  as the mixture of all task distributions $P_{D_{M_i}}$ from the environment $\tau$ averaged with $P_\tau$. From Theorem~\ref{Theorem: Meta-Gibbs}, an operational interpretation of this inequality is that for fixed meta Gibbs algorithm $P_{W_{1:m},U|D_{M_{1:m}}}^\gamma$\!, the meta generalization error will increase if we mix the datasets from different meta-training tasks, compared to treating different meta-training tasks separately.

     





To deepen our understanding of the meta Gibbs algorithm, we apply the expansion of lautum information in \cite[Eq. (52)]{palomar2008lautum} and chain rule of mutual information to Theorem~\ref{Theorem: Meta-Gibbs},
\begin{align}
    &I_{\mathrm{SKL}}(U,W_{1:m};D_{M_{1:m}}) \nn\\
    &= I_{\mathrm{SKL}}(U;D_{M_{1:m}})
       +I(W_{1:m};D_{M_{1:m}}|U) \\
       &\quad+D(P_{W_{1:m}|U} \| P_{W_{1:m}|U,M_{1:m}} |P_{U} P_{M_{1:m}} ).  \nn
\end{align}
Here, the first $I_{\mathrm{SKL}}(U;D_{M_{1:m}})$ term reflects the generalization error caused by learning the shared meta parameter $U$, and the remaining conditional information and divergence terms correspond to the generalization error in task-specific parameters.

\subsection{Example: Mean Estimation}\label{sec:mean-example}
We now generalize the mean estimation problem considered in \cite{OurGibbsPaper,bu2022characterizing} to the meta-learning setting, where the symmetrized KL information can be computed easily. Details are provided in Appendix~\ref{app: Mean Estimation}.

Consider the problem of estimating the mean $\vmu \in \mathbb{R}^d$ of the test task using samples from $m$ different meta-training tasks $D_{M_{1:m}}=\{\{Z^{M_i}_j\}_{j=1}^{n}\}_{i=1}^m$, and $D_{T}=\{Z^{T}_j\}_{j=1}^{n}$, where each task has $n$ i.i.d. samples.
We assume that the samples
from the meta-training and test tasks satisfying $\mathbb{E}[Z^{M_i}]=\vmu_{M_i}$ and $\mathrm{cov}[Z^{M_i}] = \sigma_Z^2 \mI_d$,  $\forall i\in[m]$, and $\mathbb{E}[Z^{T}]=\vmu_{T}$ and $\mathrm{cov}[Z^{T}] = \sigma_Z^2 \mI_d$, respectively. Thus, the environment $\tau$ will generate tasks with different mean $\vmu_{M_i} \sim \mathcal{N}(0, \sigma_\tau^2 \mI_d)$, but the covariance matrices of all tasks are the same.
We adopt the following regularized mean-squared loss $\ell(\vw,\vu, \vz) = \alpha \|\vz-\vw\|_2^2 +(1-\alpha)\|\vu-\vw\|_2^2$, for $\vw,\vu, \vz \in \mathbb{R}^d$, $\alpha\in[0,1]$, and assume uniform distribution over the entire space (improper prior) $\pi (\vw)$ to simplify the computation.

For this setting, the $(\gamma, \pi(\vu,\vw_{1:m}), L_E(\vu,\vw_i,d_{M_i}))$)-Gibbs algorithm is given by the Gaussian posterior distribution, $P_{W_{1:m},U|D_{M_{1:m}}}^\gamma\!\!\!\!\sim \mathcal{N}(\vmu_{W_{1:m},U},\mSigma)$, where $\vmu_{W_{1:m},U} \in \mathbb{R}^{(m+1)d}$, and 
\begin{align}
    \vmu_{W_i} = \alpha \bar{Z}^{M_i} +(1-\alpha)\bar{Z}^{M_{1:m}}, \quad
    \vmu_{U} = \bar{Z}^{M_{1:m}}.
\end{align}
Here, the notations     
\begin{align}
    \bar{Z}^{M_i} \triangleq \frac{1}{n} \sum_{j=1}^n  Z^{M_i}_j,\quad
    \bar{Z}^{M_{1:m}} \triangleq \frac{1}{m} \sum_{i=1}^m  \bar{Z}^{M_i},
\end{align}
are sample means of each meta-training task and the sample mean across all training tasks, respectively. Moreover, the covariance matrix has the following structure, 
\begin{align*}
    \mSigma^{-1}\!=\! \frac{2\gamma}{m}\!\begin{bmatrix}
\mI_d & \cdots & 0 & (\alpha-1)\mI_d\\
\vdots & \ddots & \vdots & \vdots \\
0&\cdots  & \mI_d & (\alpha-1)\mI_d\\
(\alpha-1) \mI_d &\cdots &(\alpha-1)\mI_d& m(1-\alpha)\mI_d
\end{bmatrix},
\end{align*}
which demonstrates the conditional independence between $W_i$ and $W_j$ given $U$ for any $i\ne j$. 

Since $P_{W_{1:m},U|D_{M_{1:m}}}^\gamma$ is Gaussian, the symmetrized KL information does not depend on the distribution $P_{Z}^{M_i}$ as long as $\mathrm{cov}[Z^{M_i}] = \sigma_Z^2 \mI_d$, i.e.,
\begin{align}
    I_{\mathrm{SKL}}(U,W_{1:m};D_{M_{1:m}}) = \frac{2 \gamma \alpha((m-1)\alpha +1) d\sigma_Z^2}{mn}.
\end{align}

From Theorem~\ref{Theorem: Meta-Gibbs}, the expected meta generalization error of this algorithm can be computed exactly as:
\begin{align}\label{equ:mean_meta}
    \overline{\text{gen}}(P_{W_{1:m},U|D_{M_{1:m}}}^\gamma,\tau) 
     = \frac{2 \alpha^2 d\sigma_Z^2}{n} +\frac{2\alpha(1-\alpha)d\sigma_Z^2}{mn},
\end{align}
which gives a rate of $\mathcal{O}(\frac{d}{mn}+\frac{d}{n})$.

When $\alpha=1$, the loss function $\ell(\vw,\vu, \vz) = \|\vz-\vw\|_2^2$ does not depend on the meta parameter $\vu$ anymore, which suggests no interaction between different meta-training tasks, and $U$ can be set arbitrarily. Thus, the meta generalization error in \eqref{equ:mean_meta} reduces to $\frac{2  d\sigma_Z^2}{n}$, which is precisely the generalization error of the ERM algorithm with $n$ i.i.d samples from $P_Z^{T}$ in supervised learning setting (see, \cite{OurGibbsPaper}).

When $\alpha=0$, the loss function $\ell(\vw,\vu, \vz) = \|\vu-\vw\|_2^2$ does not depend on any samples. In this case,  the meta generalization error in \eqref{equ:mean_meta}  is 0. 

For general $\alpha \in (0,1)$, it can be verified that the meta generalization error is always smaller than $\frac{2  d\sigma_Z^2}{n}$, i.e., the generalization error of ERM in supervised learning. 

\begin{remark}[Effect of $P_\tau$] As shown in \eqref{equ:mean_meta}, the meta generalization error of this mean estimation problem does not depend on the meta distribution $P_\tau$, where the variance $\sigma_\tau^2$ captures the diversity of the means $\vmu_{M_i}$ for different meta-training tasks. One reason is that the effect of the means is canceled out in meta generalization error by subtracting the empirical meta risk from the population meta risk. Although different $\sigma_\tau^2$ do not change meta generalization errors in this example, a large $\sigma_\tau^2$ implies less similarity between different tasks, and it will lead to large population meta risks. Another reason is that we set sample variance $\sigma_Z^2$ to be the same across all tasks. When environment $\tau$ generates tasks with different sample variances, meta generalization error will depend on $P_\tau$. 
\end{remark}

\section{Meta Generalization Error of the Super-task Gibbs Algorithm}

In this section, we analyze the super-task framework for meta-learning introduced in \cite{hellstrom2022evaluated} from the perspective of the Gibbs algorithm, and we offer the exact characterization of the meta generalization error.

\subsection{Notation}

We adopt the notation used in \cite{hellstrom2022evaluated} for this section. The matrix $\mZ \in \mathcal{Z}^{n \times 4 m}$ represents the entire dataset, where we divide the columns of the matrix into $2m$ groups. Each group consists of a pair of columns, where the first and second columns are the first group, the third and fourth form the second group, and so on. Each group contains $2n$ samples i.i.d. generated from the same meta task drawn from the meta distribution $P_\tau$. 
The columns in each group are labeled, with the first column labeled $0$ and the second column labeled $1$. We introduce the notation $\mZ_{j,l} \in \mathcal{Z}^{2 m}$, where $j \in [n]$ and $l \in \{0,1\}$, as a row vector formed by the $j$-th element in the column labeled by $l$ in each of the $2m$ groups. 


\begin{figure}[t!]
    \centering
    \includegraphics[scale=0.3]{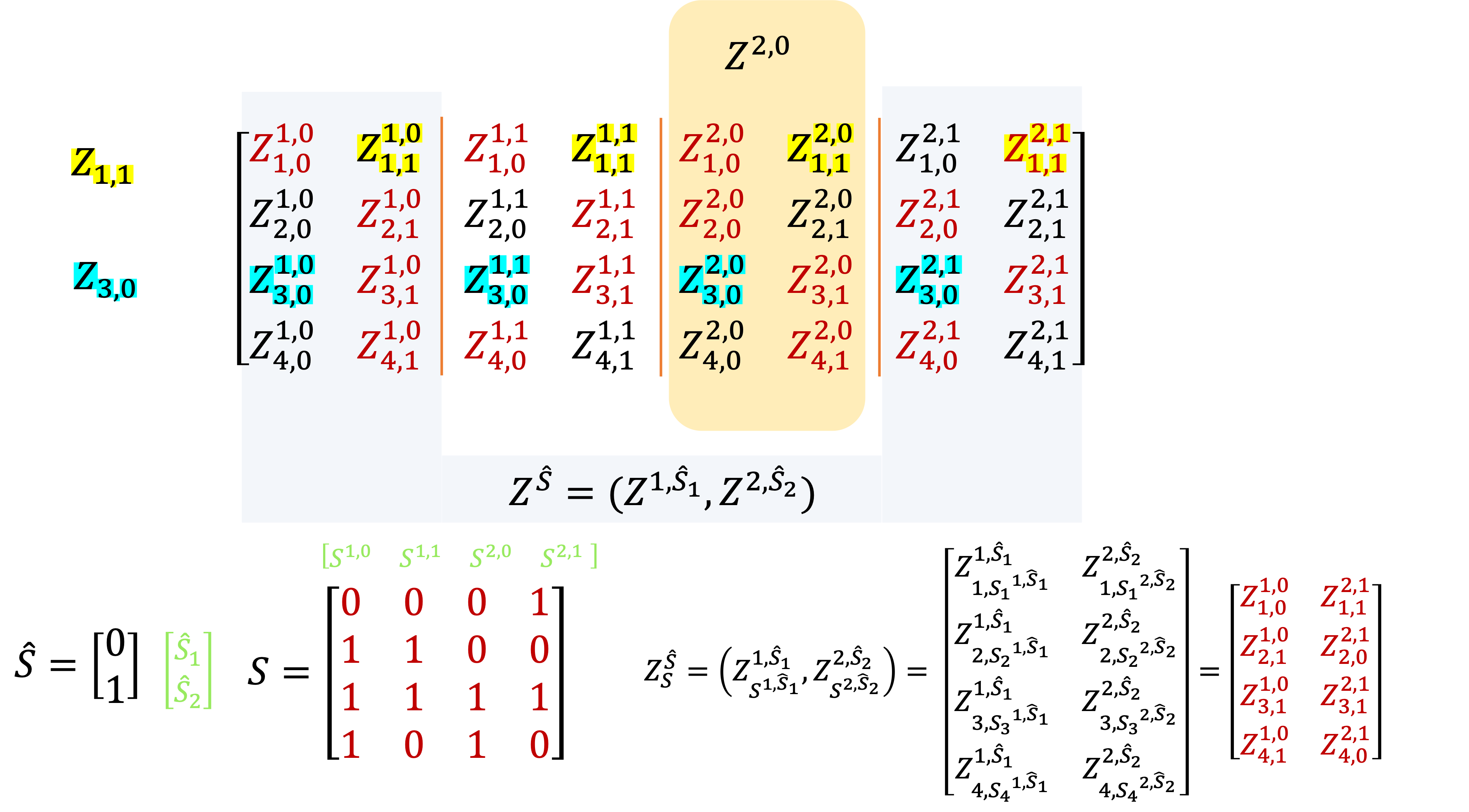}
    \vspace{-1em}
    \caption{A graphical representation of the notation system. We chose $m=2$, i.e., 4 meta tasks, and $n=4$, i.e., 8 data samples per task.}
    \label{fig:selection_of_variables}
\end{figure}

To differentiate between different meta tasks, we further label these $2m$ tasks with $(i,k)$ for $i \in [m]$ and $k \in \{0,1\}$. In addition, we use super-scripts to choose the $(i,k)$-th meta-task among the $2m$ tasks. Thus, $\mZ_{j,l}^{i,k} \in \mathcal{Z}$ is the $(i,k)$-th element of the vector $\mZ_{j,l}$.
In summary, we use superscripts to select among meta-tasks and subscripts to select among samples, as shown in Fig.~\ref{fig:selection_of_variables}.

We define a meta-training task membership vector $\hat{S} \in \{0,1\}^{m}$, where each element $\hat{S}_i $ is i.i.d. drawn from $ \textrm{Bern}(1/2)$. The meta-training tasks are selected according to the elements in $\{ (i, \hat{S}_i): i \in [m] \}$ and the meta test tasks are selected according to $\{ (i, -\hat{S}_i): i \in [m] \}$, where $-\hat{S}_i \triangleq 1- \hat{S}_i$. Within each meta task, we have $2n$ data samples, and we randomly select half of them as the training samples and the remaining as the test samples using a randomly generated matrix $S \in \{0,1\}^{n \times 2 m}$, where the elements $S^{i,k}_j$ are drawn from $\textrm{Bern}(1/2)$ for $i \in [m]$, $k \in {0,1}$, and $j \in [n]$. Each column of $S$ is a binary vector of length $n$ that indicates which sample is selected as the training data. Our complete meta-training dataset is formed by $\{ \mZ^{i,\hat{S}_i}_{j,S_j^{i,\hat{S}_i}} \}_{i,j=1}^{m,n}$.



\subsection{Characterization of Expected Meta Generalization Error }
For given membership variables $S$ and $\hat{S}$, we can rewrite the individual empirical risk for task $(i,\hat{S}_i)$ under this super-task framework as 
\begin{align}
    L_E(U,W^{i,\hat{S}_i},\mZ^{i,\hat{S}_i}_{S})\triangleq\frac{1}{n}\sum_{j=1}^{n}\ell(U,W^{i,\hat{S}_i},\mZ^{i,\hat{S}_i}_{j,S^{i,\hat{S}_i}_j}),
\end{align}
and the joint empirical risk for all meta-training tasks as
\begin{align}
    L_E(U,W^{\hat{S}},\mZ^{\hat{S}}_{S})\triangleq\frac{1}{k}\sum_{i=1}^{k}L_E(U,W^{i,\hat{S}_i},\mZ^{i,\hat{S}_i}_{S}).
\end{align}

Similar to the meta Gibbs algorithm, we can consider the super-task Gibbs algorithm for meta-training tasks using the joint empirical risk, i.e., $(\gamma,\pi(u, w^{\hat{S}}),L_E ( u, w^{\hat{S}}, \mZ^{\hat{S}}_S ))$-Gibbs algorithm
\begin{align}\label{equ:super-task-gibbs}
    P_{W^{\hat{S}},U|S,\hat{S},\mZ}^\gamma (  w^{\hat{S}},u) = \frac{\pi ( u, w^{\hat{S}} ) e^{-\gamma L_E ( u, w^{\hat{S}}, \mZ^{\hat{S}}_S )}}{V_1 \big( \mZ^{\hat{S}}_{S},\gamma \big)},
\end{align}
for those meta test tasks, the task specific weights $W^{-\hat{S}}$ are obtained by $(\gamma,\pi(w^{-\hat{S}}),L_E ( u, w^{-\hat{S}}, \mZ^{-\hat{S}}_S ))$-Gibbs algorithm for a given $U=u$,
\begin{align*}
    P_{W^{-\hat{S}}|U,S,\hat{S},\mZ}^\gamma (  w^{-\hat{S}} ) = \frac{\pi (  w^{-\hat{S}}) e^{ - \gamma L_E ( u, w^{-\hat{S}}, \mZ_{S}^{-\hat{S}}) } }{V_2 ( U, \mZ^{-\hat{S}}_S, \gamma )}.
\end{align*}

Inspired by \cite{hellstrom2022evaluated}, we define the following four different types of losses using the membership variables $S$ and $\hat{S}$.
\begin{align}
\label{eq:Lhat}\widehat{L} &= \Expt_{U,W,\mZ,S,\hat{S}} L_E \left( U,W^{\hat{S}},\mZ^{\hat{S}}_{S} \right), \\
\label{eq:Lbar}\bar{L} &= \Expt_{U,W,\mZ,S,\hat{S}} L_E \left( U,W^{\hat{S}},\mZ^{\hat{S}}_{-S} \right), \\
\label{eq:Ltilde}\widetilde{L} &= \Expt_{U,W,\mZ,S,\hat{S}} L_E \left( U,W^{-\hat{S}},\mZ^{-\hat{S}}_{S} \right),\\
\label{eq:Lp} L_{P} &= \Expt_{U,W,\mZ,S,\hat{S}} L_E \left( U,W^{-\hat{S}},\mZ^{-\hat{S}}_{-S} \right),  
\end{align}
where $\widehat{L}$ is the expected empirical meta risk evaluated on meta-training tasks, $L_{P}$ is the population meta risk evaluated on unseen tasks. The remaining two losses are the expected auxiliary test loss $\bar{L}$, which is the loss on test data for training tasks, and the expected auxiliary training loss $\widetilde{L}$, which is the loss on training data for test tasks.

The expected meta generalization error in super-task setting is 
$\overline{\text{gen}}(P_{W^{\hat{S}},U|S,\hat{S},\mZ},\tau)\triangleq 
\mathbb{E}_{P_\tau} [ L_{P} - \widehat{L}]$.

The following theorem characterizes the meta generalization error using conditional symmetrized KL information by decomposing it into these four different types of losses. The proof is provided in Appendix~\ref{app:thm2}. 
%
%
\begin{theorem}\label{Theorem:super-task}
For the super-task Gibbs algorithm defined in~\eqref{equ:super-task-gibbs}, it can be shown
\begin{enumerate}
\item $    \big( L_P + \bar{L}  + \widetilde{L} + \widehat{L}  \big) - \gamma  \widehat{L}=\cfrac{4}{\gamma}I_{SKL}  \big( U, W^{\hat{S}} ; S, \hat{S} | \mZ  \big)$,
    \item $  \big( \bar{L} - \widehat{L}  \big) = \cfrac{2}{\gamma} I_{SKL}  \big( U, W^{\hat{S}} ; S | \hat{S}, \mZ  \big)$,
     \item $  \big( \widetilde{L} - \widehat{L}  \big) = \cfrac{2}{\gamma} I_{SKL}  \big( U, W^{\hat{S}} ; \hat{S} | S, \mZ  \big)$,
     \item     $   \big( L_P - \widetilde{L}  \big)= \cfrac{2}{\gamma} I_{SKL}  \big(  W^{-\hat{S}} ; S | U, \hat{S}, \mZ  \big)$,
\end{enumerate}
and the meta generalization error is given by
\begin{align}
    &\overline{\text{gen}}(P_{W^{\hat{S}},U|S,\hat{S},\mZ},\tau)   \\
    &= \frac{2}{\gamma }\mathbb{E}_{P_\tau}\Big[I_{SKL}  \big(  W^{-\hat{S}} ; S | U,\hat{S}, \mZ  \big) + I_{SKL}  \big( U, W^{\hat{S}} ; \hat{S} | S, \mZ  \big) \Big]. \nn
\end{align}
\end{theorem}

As shown in Theorem~\ref{Theorem:super-task}, the meta generalization error can be decomposed into two symmetrized KL information terms $I_{SKL}  \big( U, W^{\hat{S}} ; \hat{S} | S, \mZ  \big)$ and $I_{SKL}  \big(  W^{-\hat{S}} ; S | U,\hat{S}, \mZ  \big)$, which represents $L_P - \widetilde{L}$ and $\widetilde{L} - \widehat{L}$, respectively.

\section{Distribution-free Upper Bound}
In this section, we present distribution-free upper bounds for the meta Gibbs algorithm and super-task Gibbs algorithm. These bounds characterize the relationship between the meta generalization error and the number of tasks $m$ and the number of samples per task $n$. It can be utilized in situations where direct computation of symmetrized KL information is challenging.

In the following Theorem, we provide the distribution-free upper bound on meta Gibbs algorithm by combining Theorem~\ref{Theorem: Meta-Gibbs} and \cite[Theorem 5.1]{chen2021generalization}. The proof is provided in Appendix~\ref{app:thm3}.
\begin{theorem}\label{Theorem: Meta-upper bound} Suppose that the meta target training samples $D_{M_i}=\{Z_j^{M_i}\}_{j=1}^n$ are i.i.d generated from the distribution $P_Z^{M_i}$, and the non-negative loss function $\ell(u,w,Z)$ is $\sigma_{\text{meta}}$-sub-Gaussian
under distribution $Z\sim P_Z^{M_i}$ and  $M_i\sim P_\tau $ for all $u \in \mathcal{U}$ and $w \in \mathcal{W}$.
If we further assume $C_{\text{meta}}\le \frac{L(U,W_{1:m};D_{M_{1:m}})}{I(U,W_{1:m};D_{M_{1:m}})}$ for some $C_{\text{meta}} \ge 0$, then for the meta Gibbs algorithm in~\eqref{Eq: meta Gibbs algorithm}, we have
\begin{equation}
    \overline{\text{gen}}(P_{W_{1:m},U|D_{M_{1:m}}}^\gamma,\tau) \le \frac{2\sigma_\text{meta}^2 \gamma}{(1+C_{\text{meta}})mn}.
\end{equation}
\end{theorem}

As shown in~\cite{xu2017information}, the sub-Gaussian condition in Theorem~\ref{Theorem: Meta-upper bound} holds for all bounded loss functions.

\begin{remark}
In comparison to the meta generalization upper bounds of the general meta learning algorithm in \cite{chen2021generalization,hellstrom2022evaluated} that scale as $\mathcal{O}(\frac{1}{\sqrt{mn}})$, we prove that the meta generalization error of meta Gibbs algorithm has a faster convergence rate $\mathcal{O}(\frac{1}{mn})$. 
\end{remark}

\begin{remark}
It can be verified easily that the loss function $\ell(\vw,\vu, \vz)$ considered in the mean estimation example  in Sec. \ref{sec:mean-example} is not bounded and does not satisfy the sub-Gaussian assumption in Theorem~\ref{Theorem: Meta-upper bound}, which results in a rate of $\mathcal{O}(\frac{1}{mn}+\frac{1}{n})$ instead of the faster rate $\mathcal{O}(\frac{1}{mn})$. 
\end{remark}

Now, we provide distribution-free upper bound on the super-task Gibbs algorithm by combining Theorem~\ref{Theorem:super-task} and \cite[Corollary~1]{hellstrom2022evaluated}. The proof is provided in Appendix~\ref{app:thm4}.
\begin{theorem}\label{Theorem: super-task-bound}
 If  the non-negative loss function is bounded, i.e., $\ell(u,w,z)\in[0,1]$, then for the super-task Gibbs algorithm defined in~\eqref{equ:super-task-gibbs}, we have
 \begin{equation}
     \overline{\text{gen}}(P_{W^{\hat{S}},U|S,\hat{S},\mZ},\tau) \le \frac{\gamma}{m}+\frac{\gamma}{n}.
 \end{equation}
\end{theorem}

\begin{remark}
Compared with the bound in Theorem~\ref{Theorem: Meta-upper bound}, the rate we obtained using super task framework is $\mathcal{O}(\frac{1}{m}+\frac{1}{n})$, which is sub-optimal. We believe this is due to the triangle inequality $L_{P} - \widehat{L}\le |L_{P} -\widetilde{L}| +|\widetilde{L}-\widehat{L}|$
used by the two-step method in \cite[Theorem 1]{hellstrom2022evaluated}, where a similar sub-optimal bound using this approach is obtained in \cite[Corollary 6]{hellstrom2022evaluated}. Although Theorem~\ref{Theorem:super-task} adopts a similar decomposition involving $\widetilde{L}$, our characterization of the meta generalization error is exact.  
\end{remark}

\section{Conclusion and Future Works}
\label{sec:conclusion}

We characterize the meta generalization error for the joint-training approach via the meta Gibbs algorithm in terms of symmetrized KL information and the super-task Gibbs algorithm in terms of conditional symmetrized KL information, respectively.
We also develop distribution-free upper bounds, which yield better estimates of the convergence rate compared to those available in the existing literature.

In future work, we plan to extend our framework to the alternate-training approach. This will include applying asymptotic analysis---similar to \cite{aminian2022information}---and provide an exact characterization in the asymptotic regime in which $\gamma\rightarrow \infty$.

\section*{Acknowledgements}
 Harsha Vardhan Tetali is supported by NSF EECS-1839704 and NSF CISE-1747783. Gholamali Aminian is supported by the UKRI Prosperity Partnership Scheme (FAIR) under the EPSRC Grant EP/V056883/1. M.~R.~D.~Rodrigues and Gholamali Aminian are also supported by the Alan Turing Institute. This work has also been supported in part by the MIT-IBM Watson AI Lab under Agreement No. W1771646, AFRL under Cooperative Agreement No.~FA8750-19-2-1000, NSF under Grant No.~CCF-1816209.



\bibliography{Ref}
\bibliographystyle{IEEEtran}



\newpage

\appendix
\subsection{Proof of Theorem~\ref{Theorem: Meta-Gibbs}} \label{app:thm1}
Recall the definition of symmetrized KL information,
\begin{align}
    &I_{\mathrm{SKL}}(U,W_{1:m};D_{M_{1:m}})
    \nn\\
    & = \mathbb{E}_{P_{W_{1:m},U,D_{M_{1:m}}}} \Big[\log\Big(\frac{P_{W_{1:m},U|D_{M_{1:m}}}^\gamma}{P_{W_{1:m},U}}\Big) \Big] \nn \\
    &\qquad- \mathbb{E}_{P_{W_{1:m},U}  P_{D_{M_{1:m}}}} \Big[\log\Big(\frac{P_{W_{1:m},U|D_{M_{1:m}}}^\gamma}{P_{W_{1:m},U}}\Big) \Big] \nn\\
    &= \mathbb{E}_{P_{W_{1:m},U,D_{M_{1:m}}}} [\log(P_{W_{1:m},U|D_{M_{1:m}}}^\gamma) ] \nn \\
    &\qquad- \mathbb{E}_{P_{W_{1:m},U}  P_{D_{M_{1:m}}}} [\log(P_{W_{1:m},U|D_{M_{1:m}}}^\gamma) ] \nn\\
    &= \mathbb{E}_{P_{W_{1:m},U,D_{M_{1:m}}}} \Big[\log\frac{\pi(U, W_{1:m}) }{V(D_{M_{1:m}},\gamma)} \Big] \nn \\
    &\qquad- \mathbb{E}_{P_{W_{1:m},U}  P_{D_{M_{1:m}}}} \Big[\log\frac{\pi(U, W_{1:m}) }{V(D_{M_{1:m}},\gamma)} \Big] \nn\\
    &\qquad +\gamma  \mathbb{E}_{P_{W_{1:m},U}  P_{D_{M_{1:m}}}} [ L_E(U,W_{1:m},D_{M_{1:m}}) ]\nn\\
    &\qquad -\gamma \mathbb{E}_{P_{W_{1:m},U,D_{M_{1:m}}}} [  L_E(U,W_{1:m},D_{M_{1:m}})]\nn \\
    &= \gamma  \mathbb{E}_{P_{W_{1:m},U}  P_{D_{M_{1:m}}}} [ L_E(U,W_{1:m},D_{M_{1:m}}) ]\nn\\
    &\qquad -\gamma \mathbb{E}_{P_{W_{1:m},U,D_{M_{1:m}}}} [  L_E(U,W_{1:m},D_{M_{1:m}})].
\end{align}

For the second term, we have
\begin{align}
    &\mathbb{E}_{P_\tau}\big[\mathbb{E}_{P_{W_{1:m},U,D_{M_{1:m}}}} [ L_E(U,W_{1:m},D_{M_{1:m}}) ]\big]\nn \\
    & =  \mathbb{E}_{P_\tau}\big[\mathbb{E}_{P_{U,D_{M_{1:m}}}}   [\frac{1}{m}\sum_{i=1}^m \mathbb{E}_{P_{W_i|U,D_{M_i} }}[ L_E(U,W_{i},D_{M_{i}}) ]]\big]\nn \\
    & = \mathbb{E}_{P_\tau}\big[\mathbb{E}_{P_{U,D_{M_{1:m}}}}[L_E (U,D_{M_{1:m}})]\big],
\end{align}
which corresponds to the expected empirical meta risk. We need to show that the first term is the population meta risk, 
\begin{align}
    &\mathbb{E}_{P_\tau}\big[\mathbb{E}_{P_{W_{1:m},U}  P_{D_{M_{1:m}}}} [ L_E(U,W_{1:m},D_{M_{1:m}}) ]\big]\nn\\
    & =  \mathbb{E}_{P_\tau}\big[\mathbb{E}_{P_{W_{1:m},U}}   [\frac{1}{m}\sum_{i=1}^m \mathbb{E}_{P_{D_{M_i} }}[ L_E(U,W_{i},D_{M_{i}}) ]]\big]\nn \\
    & = \mathbb{E}_{P_\tau}\big[\mathbb{E}_{P_{W_{1:m},U}}   [\frac{1}{m}\sum_{i=1}^m  L_P(U,W_i,P_{D_{M_i}})]\big]\nn \\
     & = \mathbb{E}_{P_\tau}\big[\mathbb{E}_{P_{U}}   [\frac{1}{m}\sum_{i=1}^m \mathbb{E}_{P_{W_i|U}}[ L_P(U,W_i,P_{D_{M_i}})]]\big]\nn \\
  & = \mathbb{E}_{P_\tau}\Big[\mathbb{E}_{P_{U}}   \big[\frac{1}{m}\sum_{i=1}^m\mathbb{E}_{P_{D_{M_i}}}[ \mathbb{E}_{P_{W_i|U,D_{M_i}}}[ L_P(U,W_i,P_{D_{M_i}})]]\big]\Big]\nn \\
    & = \mathbb{E}_{P_\tau}\Big[\mathbb{E}_{P_{U}}   \big[\frac{1}{m}\sum_{i=1}^m\mathbb{E}_{P_{D_{T}}}[ \mathbb{E}_{P_{W_i|U,D_{T}}}[ L_P(U,W_i,P_{D_{T}})]]\big]\Big]\nn \\
    & \overset{(a)}{=} \mathbb{E}_{P_{U}}   \Big[\mathbb{E}_{P_\tau}\big[ \mathbb{E}_{P_{D_{T}}}[ \mathbb{E}_{P_{W_T|U,D_{T}}}[ L_P(U,W_T,P_{D_{T}})]]\big]\Big]\nn \\
     & = \mathbb{E}_{P_{U}}[L_P (U,\tau)],
\end{align}
where (a) holds as both $T$ and $M_i$ are generated independently from the same distribution $P_\tau$.

\subsection{Example: Mean Estimation}\label{app: Mean Estimation}
The following lemma from \cite{palomar2008lautum} characterizes the mutual and lautum information for the Gaussian channel.

\begin{lemma}{\cite[Theorem 14]{palomar2008lautum}}\label{lemma:Gaussian}
Consider the following model
\begin{equation}
    \mY = \mA \mX+\mN_{\mathrm{G}},
\end{equation}
where $\mX \in \mathbb{R}^{d_X}$ denotes the input
random vector with zero mean (not necessarily
Gaussian), $\mA \in \mathbb{R}^{d_Y \times d_X}$ denotes the linear transformation undergone by the input, $\mY\in \mathbb{R}^{d_Y}$ is the
output vector, and $\mN_{\mathrm{G}}\in \mathbb{R}^{d_Y}$ is a
Gaussian noise vector independent of $\mX$. The input and the
noise covariance matrices are given by
$\mSigma$ and $\mSigma_{N_{\mathrm{G}}}$.
Then, we have
\begin{align}
    I(\mX;\mY) &= \frac{1}{2}\mathrm{tr}\big(\mSigma_{N_{\mathrm{G}}}^{-1} \mA \mSigma \mA^\top \big) - D\big(P_\mY\|P_{N_{\mathrm{G}}} \big),  \\
    L(\mX;\mY) &= \frac{1}{2}\mathrm{tr}\big(\mSigma_{N_{\mathrm{G}}}^{-1} \mA \mSigma \mA^\top \big) + D\big(P_\mY\|P_{N_{\mathrm{G}}}).
\end{align}
\end{lemma}
In the meta Gibbs algorithm, the task-specific parameter $W_{1:m}$ and the meta parameter $U$ can be written as
\begin{align}\label{equ:Gaussian_meta}
        W_i& = \alpha \bar{Z}^{M_i} +(1-\alpha)\bar{Z}^{M_{1:m}} +N_{i} \nn \\
        &=\frac{\alpha}{n}\sum_{j=1}^n (Z^{M_i}_j-\vmu_i) + \frac{1-\alpha}{mn}\sum_{k=1}^m \sum_{j=1}^n (Z^{M_k}_j- \vmu_k) \nn \\
        &\qquad + \alpha\vmu_i + \frac{1-\alpha}{m}\sum_{k=1}^m\vmu_k +N_{i}, 
        \\
        U& = \bar{Z}^{M_{1:m}}+N_{U}\nn \\
        &=\frac{1}{mn} \sum_{i=1}^m \sum_{j=1}^n (Z^{M_i}_j- \vmu_i)+\frac{1}{m}\sum_{i=1}^m\vmu_i+N_{U},
\end{align}
where the additive Gaussian noise $N$ is zero mean and has the covariance 
\begin{align}
    \mSigma_N^{-1}\!=\! \frac{2\gamma}{m}\!\begin{bmatrix}
\mI_d & \cdots & 0 & (\alpha-1)\mI_d\\
\vdots & \ddots & \vdots & \vdots \\
0&\cdots  & \mI_d & (\alpha-1)\mI_d\\
(\alpha-1) \mI_d\!&\cdots &(\alpha-1)\mI_d\!& m(1-\alpha)\mI_d
\end{bmatrix},
\end{align}

Thus, for fixed meta-training samples $d_{M_{1:m}}$, we can set $P_{N_{\mathrm{G}}}$ with $\mSigma_{N_\mathrm{G}} = \mSigma_{N}$ and $\mSigma = \sigma_Z^2 I_{mnd}$ in Lemma~\ref{lemma:Gaussian}, which gives
\begin{align}\label{equ:trace}
    I_{\mathrm{SKL}}(W_{1:m},U;D_{M_{1:m}}) & = \mathrm{tr}\big(\mSigma_{N_{\mathrm{G}}}^{-1} \mA \mSigma \mA^\top \big) \nn \\
    & = \mathrm{tr}\big(\sigma_Z^2 \mSigma_{N_{\mathrm{G}}}^{-1} \mA  \mA^\top \big).
\end{align}
From~\eqref{equ:Gaussian_meta}, for $\mA \in \mathbb{R}^{(m+1)d \times mnd}$, we can obtain that
\begin{align}
    \mA  \mA^\top = \begin{bmatrix}
 \mB & \cdots & \frac{1}{mn}\\
\vdots & \ddots & \vdots\\
\frac{1}{mn}&\cdots& \frac{1}{mn}
\end{bmatrix},
\end{align}
where $\mB \in \mathbb{R}^{md\times md}$, with all diagonal elements equal to $\frac{(m\alpha+(1-\alpha))^2+(m-1)(1-\alpha)^2}{m^2n}$, and all off-diagonal element equal to $\frac{(2m\alpha(1-\alpha))+m(1-\alpha)^2}{m^2n}$.

Thus, from~\eqref{equ:trace}, it can be shown that
\begin{align}
    I_{\mathrm{SKL}}(W_{1:m},U;D_{M_{1:m}}) &= \frac{2 \gamma \alpha((m-1)\alpha +1) d\sigma_Z^2}{mn}.
\end{align}



\subsection{Proof of Theorem~\ref{Theorem:super-task}}
\label{app:thm2}
We will start with the proof of 1), which connects the conditional symmetrized KL information with four different types of losses.

We note that
\begin{align}
   & I_{SKL} \left( U,W^{\hat{S}} ; S, \hat{S} |\mZ \right) \nn \\
   &=  \Expt_{P_{\mZ}} \left[  \Expt_{P_{U,W^{\hat{S}},S,\hat{S}|\mZ}} \left[ \log \left( P_{U,W^{\hat{S}},S,\hat{S}|\mZ} \right) \right] \right] \nn\\
    &\quad - \Expt_{P_{\mZ}} \left[  \Expt_{P_{U,W^{\hat{S}}|\mZ}P_{S,\hat{S}|\mZ}} \left[ \log \left(  P_{U,W^{\hat{S}},S,\hat{S}|\mZ}\right) \right] \right]. 
\end{align}
As the quantity inside the expectation does not depend on $W^{-\hat{S}}$, we have
    \begin{align}
    & I_{SKL} \left( U,W^{\hat{S}} ; S, \hat{S} |\mZ \right) \nn \\
 &=  \Expt_{P_{\mZ}} \left[  \Expt_{P_{U,W,S,\hat{S}|\mZ}} \left[ \log \left( P_{U,W^{\hat{S}}|S,\hat{S},\mZ} \right) \right] \right] \nn \\
    &\quad- \Expt_{P_{\mZ}} \left[  \Expt_{P_{U,W|\mZ}P_{S,\hat{S}|\mZ}} \left[ \log \left(  P_{U,W^{\hat{S}}|S,\hat{S},\mZ}\right) \right] \right] \nn \\
     &=  \Expt_{P_{\mZ}} \left[  \Expt_{P_{U,W,S,\hat{S}|\mZ}} \left[ \log \left( P_{U,W^{\hat{S}}|S,\hat{S},\mZ} \right) \right] \right] \nn \\
    &\quad- \Expt_{P_{\mZ}} \left[  \Expt_{P_{U,W|\mZ}P_{S,\hat{S}}} \left[ \log \left(  P_{U,W^{\hat{S}}|S,\hat{S},\mZ}\right) \right] \right] \nn \\
      &=  \gamma \Expt_{P_{\mZ}} \left[  \Expt_{P_{U,W|\mZ}P_{S',\hat{S}'}} \left[ L_E \left( U, W^{\hat{S}}, \mZ_{S'}^{\hat{S}'} \right) \right] \right]  \nn \\ 
    &\quad-  \gamma \Expt_{P_{\mZ}} \left[  \Expt_{P_{U,W,S,\hat{S}|\mZ}} \left[ L_E \left( U, W^{\hat{S}}, \mZ_{S}^{\hat{S}} \right) \right] \right] \label{eq:meta_super_1},
\end{align}
where $S'$ is an independent copy of $S$ and $\hat{S}'$ is an independent copy of $\hat{S}$. Thus,
\begin{align*}
    & \Expt_{P_{\mZ}} \Big[  \Expt_{P_{U,W|\mZ}P_{S',\hat{S}'}} \Big[ L_E \Big( U, W^{\hat{S}}, \mZ_{S'}^{\hat{S}'} \Big) \Big] \Big] \\
    &= \Expt_{P_{\mZ}} \Big[  \Expt_{P_{U,W|\mZ}P_{S',\hat{S}'}} \Big[ \frac{1}{m n}\sum_{i=1}^k\sum_{j=1}^{n}\ell \Big(U,W^{i,\hat{S}_i},\mZ^{i,\hat{S}_i}_{j,S'_j}\Big)  \Big] \Big] \\
    &= \Expt_{P_{\mZ}} \Big[  \Expt_{P_{U,W|\mZ}} \Big[ \frac{1}{4m n}\sum_{i=1}^k\sum_{j=1}^{n}\ell \Big(U,W^{i,0},\mZ^{i,0}_{j,0}\Big)  \Big] \Big] \\
    &\quad+ \Expt_{P_{\mZ}} \Big[  \Expt_{P_{U,W|\mZ}} \Big[ \frac{1}{4m n}\sum_{i=1}^k\sum_{j=1}^{n}\ell \Big(U,W^{i,1},\mZ^{i,0}_{j,1}\Big)  \Big] \Big] \\
    &\quad+\Expt_{P_{\mZ}} \Big[  \Expt_{P_{U,W|\mZ}} \Big[ \frac{1}{4m n}\sum_{i=1}^k\sum_{j=1}^{n}\ell \Big(U,W^{i,0},\mZ^{i,1}_{j,0}\Big)  \Big] \Big] \\
    &\quad+\Expt_{P_{\mZ}} \Big[  \Expt_{P_{U,W|\mZ}} \Big[ \frac{1}{4m n}\sum_{i=1}^k\sum_{j=1}^{n}\ell \Big(U,W^{i,1},\mZ^{i,1}_{j,1}\Big)  \Big] \Big] \\
     &= \Expt_{P_{\mZ,S,\hat{S}}} \Big[  \Expt_{P_{U,W|\mZ,S,\hat{S}}} \Big[ \frac{1}{4m n}\sum_{i=1}^k\sum_{j=1}^{n}\ell \Big(U,W^{i,0},\mZ^{i,0}_{j,0}\Big)  \Big] \Big] \\
    &\quad+ \Expt_{P_{\mZ,S,\hat{S}}} \Big[  \Expt_{P_{U,W|\mZ,S,\hat{S}}} \Big[ \frac{1}{4m n}\sum_{i=1}^k\sum_{j=1}^{n}\ell \Big(U,W^{i,1},\mZ^{i,0}_{j,1}\Big)  \Big] \Big] \\
    &\quad+\Expt_{P_{\mZ,S,\hat{S}}} \Big[  \Expt_{P_{U,W|\mZ,S,\hat{S}}} \Big[ \frac{1}{4m n}\sum_{i=1}^k\sum_{j=1}^{n}\ell \Big(U,W^{i,0},\mZ^{i,1}_{j,0}\Big)  \Big] \Big] \\
    &\quad+\Expt_{P_{\mZ,S,\hat{S}}} \Big[  \Expt_{P_{U,W|\mZ,S,\hat{S}}} \Big[ \frac{1}{4m n}\sum_{i=1}^k\sum_{j=1}^{n}\ell \Big(U,W^{i,1},\mZ^{i,1}_{j,1}\Big)  \Big] \Big] \\
    &= \Expt_{P_{\mZ,S,\hat{S}}} \Big[  \Expt_{P_{U,W|\mZ,S,\hat{S}}} \Big[ \frac{1}{4}  L_E \Big(U,W^{\hat{S}},\mZ_{S}^{\hat{S}}\Big) \Big] \Big] \\
    &\quad+ \Expt_{P_{\mZ,S,\hat{S}}} \Big[  \Expt_{P_{U,W|\mZ,S,\hat{S}}} \Big[ \frac{1}{4}  L_E \Big(U,W^{\hat{S}},\mZ_{-S}^{\hat{S}}\Big) \Big] \Big] \\
     &\quad+ \Expt_{P_{\mZ,S,\hat{S}}} \Big[  \Expt_{P_{U,W|\mZ,S,\hat{S}}} \Big[ \frac{1}{4}  L_E \Big(U,W^{-\hat{S}},\mZ_{S}^{-\hat{S}}\Big) \Big] \Big] \\
     &\quad+ \Expt_{P_{\mZ,S,\hat{S}}} \Big[  \Expt_{P_{U,W|\mZ,S,\hat{S}}} \Big[ \frac{1}{4}  L_E \Big(U,W^{-\hat{S}},\mZ_{-S}^{-\hat{S}}\Big) \Big] \Big],
\end{align*}
where the last step follows from the fact that each element of $(S,\hat{S})$ is one of $(0,0),(0,1),(1,0)$ or $(1,1)$.
Finally using \eqref{eq:Lhat}, \eqref{eq:Lbar}, \eqref{eq:Ltilde} and \eqref{eq:Lp}, we get, \begin{align*}
    & \Expt_{P_{\mZ}} \left[  \Expt_{P_{U,W|\mZ}P_{S',\hat{S}'}} \left[ L_E \left( U, W^{\hat{S}}, \mZ_{S'}^{\hat{S}'} \right) \right] \right] \\
    &= \cfrac{1}{4} \left[ \widehat{L} + \bar{L} + \widetilde{L} + L_P \right].
\end{align*}
Substituting the above into \eqref{eq:meta_super_1}, and identifying that the second term in \eqref{eq:meta_super_1} corresponds to $\gamma \widehat{L}$, we have,
\begin{align}
     I_{SKL} \left( U,W^{\hat{S}} ; S, \hat{S} |\mZ \right) = \cfrac{\gamma}{4} \left[ \widehat{L} + \bar{L} + \widetilde{L} + L_P \right] - \gamma \widehat{L}.
\end{align}

Using a similar approach, we can also prove 2) and 3),
\begin{align}
    I_{SKL}  \big( U, W^{\hat{S}} ; S | \hat{S}, \mZ  \big) &= \frac{\gamma}{2}  \big( \bar{L} - \widehat{L}  \big), \\
    I_{SKL}  \big( U, W^{\hat{S}} ; \hat{S} | S, \mZ  \big) &= \frac{\gamma}{2} \big( \widetilde{L} - \widehat{L} \big) .\label{equ:res3}
\end{align}

For the proof of 4) on $I_{SKL}\left( W^{-\hat{S}}; S | U, \hat{S}, \mZ \right)$, we have
\begin{align*}
    &I_{SKL}\left( W^{-\hat{S}}; S | U, \hat{S}, \mZ \right) \\
    &= \Expt_{P_{U,\hat{S},\mZ}} \left[ \Expt_{P_{W^{-\hat{S}},S|U,\hat{S},\mZ}} \left[ \log P_{W^{\hat{S}},S|U,\hat{S},\mZ} \right]  \right] \\
    &\quad-\Expt_{P_{W^{-\hat{S}}|U,\hat{S},\mZ}P_{S|U,\hat{S},\mZ}} \left[ \log P_{W^{-\hat{S}}|S,U,\hat{S},\mZ} \right]\\
    &= \Expt_{P_{U,\hat{S},\mZ}}  \left[ \gamma  \Expt_{P_{W^{-\hat{S}}|U,\hat{S},\mZ}P_{S|U,\hat{S},\mZ}} \left[ L_{E} \left( U,W^{-\hat{S}},\mZ^{-\hat{S}}_S \right)  \right]\right] \\
    &\quad-\Expt_{P_{U,\hat{S},\mZ}}  \left[ \gamma \Expt_{P_{W^{-\hat{S}},S|U,\hat{S},\mZ}}\left[ 
L_E \left( U, W^{-\hat{S}},\mZ^{-\hat{S}}_{S} \right) \right] \right].
\end{align*}
It can be seen that the second term equals $-\gamma \widetilde{L}$, i.e., the loss on training data $S$ for test tasks $-\hat{S}$. In addition, $P_{S|U,\hat{S},\mZ} = P_S$ as $S$ is independent with $U,\hat{S}$ and $\mZ$. Thus,
\begin{align*}
    &I_{SKL}\left( W^{-\hat{S}}; S | U, \hat{S}, \mZ \right) \\
    &= \gamma \Expt_{P_{U,\hat{S},\mZ}} \left[ \Expt_{P_{W^{-\hat{S}}|U,\hat{S},\mZ} P_S} \left[L_{E}\left(U,W^{-\hat{S}},\mZ^{-\hat{S}}_S\right) \right]\right] -\gamma \widetilde{L}.
\end{align*}
Following the same argument as in the previous proof, we have that,
\begin{align*}
  &I_{SKL}\left( W^{-\hat{S}}; S | U, \hat{S}, \mZ \right) \\
  &= \frac{\gamma}{2} \Expt_{P_{U,\hat{S},\mZ} \Expt_{P_{W^{-\hat{S}},S|U,\hat{S},\mZ}}} \left[ L_{E}\left(U,W^{-\hat{S}},\mZ^{-\hat{S}}_S\right)  \right] \\
  &\quad+\frac{\gamma}{2} \Expt_{P_{U,\hat{S},\mZ} \Expt_{P_{W^{-\hat{S}},S|U,\hat{S},\mZ}}} \left[ L_{E}\left(U,W^{-\hat{S}},\mZ^{-\hat{S}}_{-S}\right)  \right] \\
  &\quad-\gamma \widetilde{L}.
\end{align*}
Finally, we have,
\begin{align}\label{equ:res4}
    I_{SKL}\left( W^{-\hat{S}}; S | U, \hat{S}, \mZ \right) = \frac{\gamma}{2} \left( L_P - \widetilde{L} \right).
\end{align}
The meta generalization error can be written as
\begin{align}
    \overline{\text{gen}}(P_{W^{\hat{S}},U|S,\hat{S},\mZ},\tau) &= \mathbb{E}_{P_\tau}[L_{P} - \widehat{L}] \nn \\
    &= \mathbb{E}_{P_\tau}[L_{P} -\widetilde{L}+\widetilde{L}-\widehat{L}],
\end{align}
combining the above equation with~\eqref{equ:res3} and~\eqref{equ:res4} completes the proof.

\subsection{Proof of Theorem~\ref{Theorem: Meta-upper bound}}\label{app:thm3}
From \cite[Theorem 5.1]{chen2021generalization}, under the sub-Gaussian condition, it has been shown that
\begin{align}\label{Eq: meta const}
   &|\overline{\text{gen}}(P_{W_{1:m}|U,D_{M_{1:m}}},P_{U|D_{M_{1:m}}},\tau)|\nn \\
&\qquad \leq
   \sqrt{\frac{2 \sigma_\text{meta}^2}{mn} \mathbb{E}_{P_\tau}[I(U,W_{1:m};D_{M_{1:m}})}].
\end{align}

Combining with Theorem~\ref{Theorem: Meta-Gibbs}, we have
\begin{align}
   &\overline{\text{gen}}(P_{W_{1:m}|U,D_{M_{1:m}}},P_{U|D_{M_{1:m}}},\tau)\nn\\
   &=\frac{ \mathbb{E}_{P_\tau}[I_{\mathrm{SKL}}(U,W_{1:m};D_{M_{1:m}})]}{\gamma}\nn \\
   &\leq \sqrt{\frac{2 \sigma_\text{meta}^2}{mn} \mathbb{E}_{P_\tau}[I(U,W_{1:m};D_{M_{1:m}})}].
\end{align}
As $I(U,W_{1:m};D_{M_{1:m}})(1+C_\text{meta})\leq I_{\mathrm{SKL}}(U,W_{1:m};D_{M_{1:m}})$ by the assumption, we have
\begin{align}
    &\frac{(1+C_{\text{meta}})}{\gamma}\mathbb{E}_{P_\tau}[I(U,W_{1:m};D_{M_{1:m}})]\nn \\
    &\le \sqrt{\frac{2\sigma_\text{meta}^2 \mathbb{E}_{P_\tau}[I(U,W_{1:m};D_{M_{1:m}})]}{mn}},
\end{align}
which implies that
\begin{equation}\label{Eq: MI upper bound meta}
    \mathbb{E}_{P_\tau}[I(U,W_{1:m};D_{M_{1:m}})]\leq \frac{2\sigma_\text{meta}^2\gamma^2}{(1+C_{\text{meta}})^2mn}.
\end{equation}
Combining \eqref{Eq: MI upper bound meta} with \eqref{Eq: meta const} completes the proof.

\subsection{Proof of Theorem~\ref{Theorem: super-task-bound}}\label{app:thm4}
In the proof of \cite[Corollary 1]{hellstrom2022evaluated}, it is shown that
\begin{align}
    | \widetilde{L} - \widehat{L} | &\leq \sqrt{\frac{2I(U;\hat{S}|\mZ,S)}{m}}, \\
    | L_P -\widetilde{L} | &\leq \sqrt{\frac{2I(W^{i,-\hat{S}_i};S^{i,-\hat{S}_i}|\mZ,\hat{S}_i)}{n}}. 
\end{align}
The first bound can be further relaxed as
\begin{align}
    | \widetilde{L} - \widehat{L} | 
    \leq \sqrt{\frac{2I(U;\hat{S}|\mZ,S)}{m}} 
    \leq \sqrt{\frac{2I(U,W^{\hat{S}};\hat{S}|\mZ,S)}{m}}, 
\end{align}
where the second inequality is due to the fact that more variables will increase mutual information.

The second bound can be upper bounded as
\begin{align}
    | L_P -\widetilde{L} | &\leq \sqrt{\frac{2I(W^{i,-\hat{S}_i};S^{i,-\hat{S}_i}|\mZ,\hat{S}_i)}{n}} \nn \\
    &\overset{(a)}{\leq} \sqrt{\frac{2I(W^{i,-\hat{S}_i};S^{i,-\hat{S}_i}|U,\mZ,\hat{S}_i)}{n}} \nn \\
    &\leq \sqrt{\frac{2I(W^{-\hat{S}};S|U,\mZ,\hat{S})}{n}}, 
\end{align}
where (a) is due to the fact that conditioning on independent variables will increase mutual information. 
Here, conditioning on $\mZ$ and $\hat{S}_i$, $U$ is independent of $S^{i,-\hat{S}_i}$, since $U$ is learned using only $\mZ^{\hat{S}}_{S^{\hat{S}}}$, not those samples indexed by $S^{-\hat{S}}$.

From Theorem~\ref{Theorem:super-task}, we have
\begin{align}
    I_{SKL}  \big( U, W^{\hat{S}} ; \hat{S} | S, \mZ  \big) &= \cfrac{\gamma}{2}  \big( \widetilde{L} - \widehat{L}  \big),  \\
    I_{SKL}\big( W^{-\hat{S}}; S | U, \hat{S}, \mZ \big) &= \frac{\gamma}{2} \big( L_P - \widetilde{L} \big).
\end{align}


Thus, we have
\begin{align}
    |\widetilde{L} - \widehat{L} | &\leq \sqrt{\frac{2I(U,W^{\hat{S}};\hat{S}|\mZ,S)}{m}} \nn \\
    & \le \sqrt{\frac{2I_{SKL}  \big( U, W^{\hat{S}} ; \hat{S} | S, \mZ  \big) }{m}}\nn \\
    &\leq  \sqrt{\frac{\gamma}{m} (\widetilde{L} - \widehat{L})}, 
\end{align}
which implies $|\widetilde{L} - \widehat{L} |  \le \frac{\gamma}{m}$, and
\begin{align}
    |L_p - \widetilde{L} | &\leq \sqrt{\frac{2 I (W^{-\hat{S}},S|U,\mZ,\hat{S})}{n}} \nn \\
    & \le \sqrt{\frac{2I_{SKL}\big( W^{-\hat{S}}; S | U, \hat{S}, \mZ \big)}{n}}\nn \\
    &\leq  \sqrt{\frac{\gamma}{n} (L_p - \widetilde{L})},
\end{align}
which gives $|L_p - \widetilde{L} |  \le \frac{\gamma}{n}$. Combining these two inequalities, we have
\begin{align}
    | L_p - \widehat{L} | \leq \gamma \left( \frac{1}{n} + \frac{1}{m} \right).
\end{align}



\end{document}